\documentclass[preprint,12pt,authoryear]{elsarticle}

\usepackage{amssymb}
\usepackage{graphicx}
\usepackage{soul}
\usepackage{subcaption}
\usepackage{float}
\usepackage{booktabs}
\usepackage{multirow}
\usepackage{moresize}
\usepackage{tabularx}
\usepackage{tabularray}
\usepackage{lineno}
\usepackage[english]{babel}

\journal{Computers and Electronics in Agriculture}

\begin{document}

\begin{frontmatter}



\title{M18K: A Comprehensive RGB-D Dataset and Benchmark for
Mushroom Detection and Instance
Segmentation}

\author[c]{Abdollah Zakeri}
\author[t]{Mulham Fawakherji}
\author[t]{Jiming Kang}
\author[t]{Bikram Koirala}
\author[t]{Venkatesh Balan}
\author[t]{Weihang Zhu}
\author[t]{Driss Benhaddou}
\author[c,t]{Fatima A. Merchant}
\affiliation[c]{organization={Department of Computer Science},
            addressline={University of Houston},
            city={Houston},
            state={TX},
            country={US}}
\affiliation[t]{organization={Department of Engineering Technology},
            addressline={University of Houston},
            city={Houston},
            state={TX},
            country={US}}



\begin{abstract}
  Automating agricultural processes holds significant promise for enhancing efficiency and sustainability in various farming practices. This paper contributes to the automation of agricultural processes by providing a dedicated mushroom detection dataset related to automated harvesting, growth monitoring, and quality control of the button mushroom produced using Agaricus Bisporus fungus. With over 18,000 mushroom instances in 423 RGB-D image pairs taken with an Intel RealSense D405 camera, it fills the gap in mushroom-specific datasets and serves as a benchmark for detection and instance segmentation algorithms in smart mushroom agriculture. The dataset, featuring realistic growth environment scenarios with comprehensive annotations, is assessed using advanced detection and instance segmentation algorithms. The paper details the dataset’s characteristics, evaluates algorithmic performance, and for broader applicability, we have made all resources publicly available including images, codes, and trained models via our GitHub repository \textit{https://github.com/abdollahzakeri/m18k}
\end{abstract}


\end{frontmatter}



\section{Introduction}
\label{sec:intro}

The integration of Artificial Intelligence (AI)-driven automation into agricultural processes has gained significant momentum, offering opportunities for increased efficiency and sustainability in various farming practices \cite{9780790,baisa_mushrooms_2022,arjun_non-destructive_2022,Nguyen_2023_building}. Among these, the automatic harvesting of edible mushrooms in large farms has emerged as an area with substantial potential for streamlining operations and reducing labor-intensive tasks.

The development of accurate and robust mushroom detection methods is essential for the success of mushroom harvesting systems, aiming to identify and locate individual mushrooms within complex and densely populated environments. While computer vision and object detection technologies have seen significant advancements, a major challenge remains due to the lack of publicly available datasets specifically tailored for mushroom detection. This impediment hinders systematic evaluation and comparison of different mushroom detection algorithms and methodologies. To address this, synthetic mushroom datasets have been introduced \cite{anagnostopoulou_realistic_2023, 9780790}, serving as a valuable tool for pre-training models. However, these datasets fall short in capturing the full breadth of challenges encountered in real-world scenarios. They often fail to accurately represent various lighting conditions, the diversity of mushrooms at different stages of growth, and the wide range of backgrounds and orientations that mushrooms can be found in. Moreover, issues such as densely populated areas, various occlusions of mushrooms, and different soil patterns are not comprehensively covered, which means that, although useful for the initial stages of model training, these datasets do not provide sufficient robustness for detection models in real-world applications. Consequently, for a detection model to be truly effective in practical scenarios, further fine-tuning on a dataset of real images is essential, ensuring the model can navigate the complexities and variability of natural environments. To address this research gap, we present a comprehensive dataset of mushroom images specifically designed to serve as a benchmark for a valid comparison of mushroom detection methods and to improve the quality of mushroom localization in images. Our dataset consists of 423 RGB-D images, captured using an Intel RealSense D405 RGB-D camera. Both the RGB and depth images have a resolution of 1,280x720 pixels, ensuring consistent image quality and compatibility across the dataset.

\begin{figure*}[!ht]
    \centering
    \begin{subfigure}{0.3\textwidth}
        \centering
        \includegraphics[width=\linewidth]{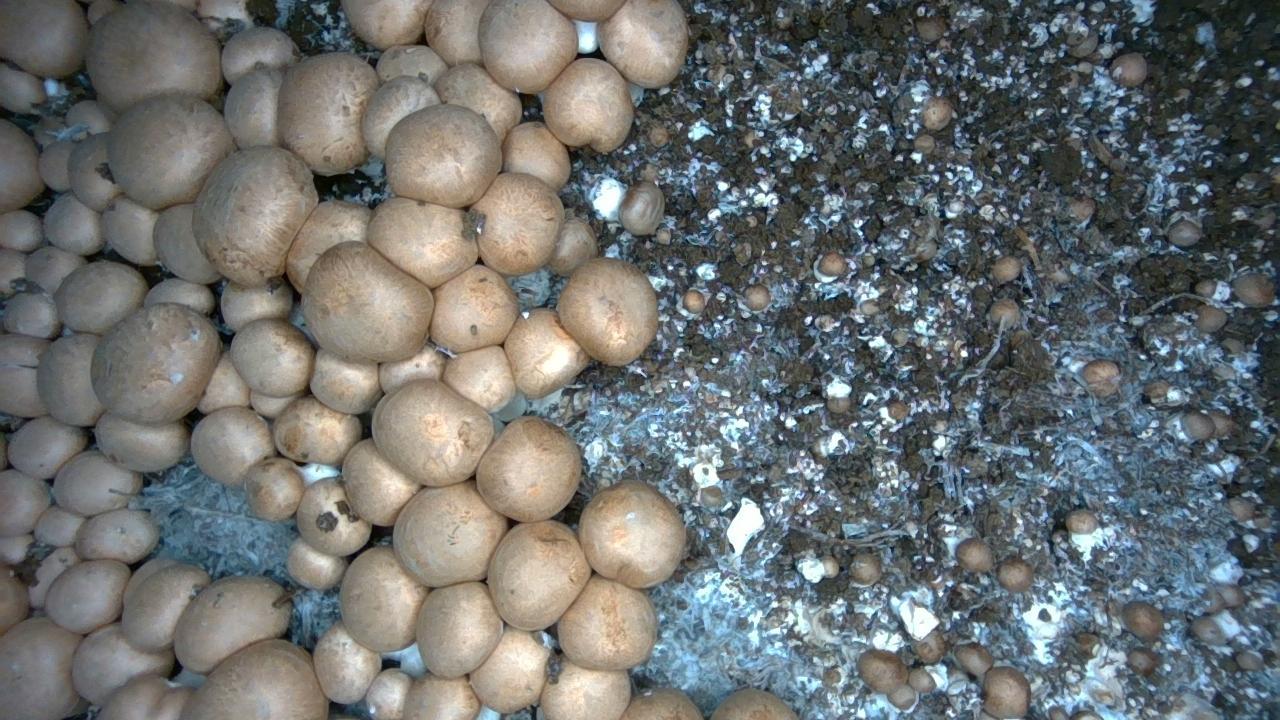}
        \caption{BB Sample Image}
        \label{fig:suba}
    \end{subfigure}
    \hfill 
    \begin{subfigure}{0.3\textwidth}
        \centering
        \includegraphics[width=\linewidth]{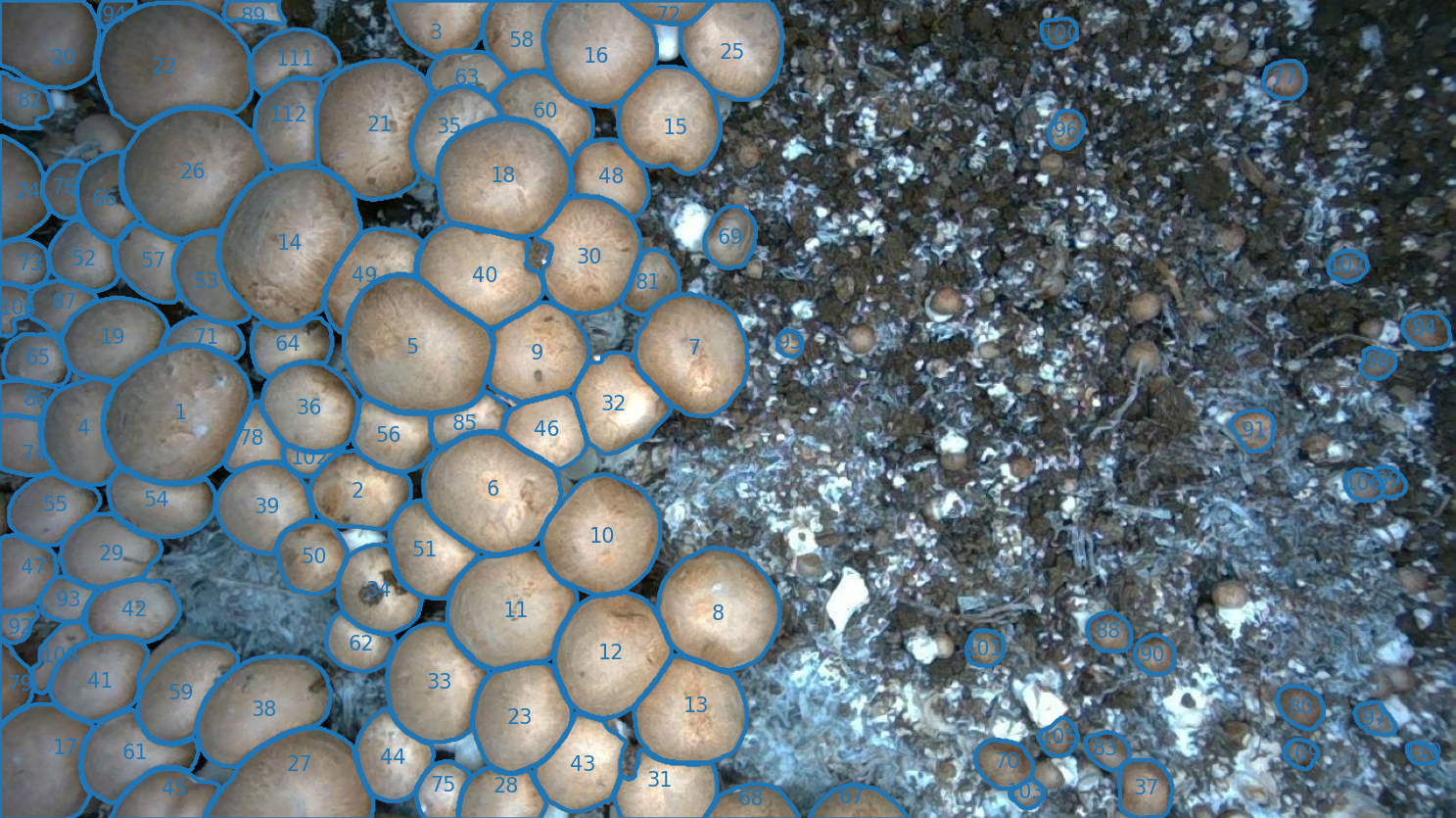}
        \caption{BB Sample Ground Truth}
        \label{fig:subc}
    \end{subfigure}
    \hfill 
    \begin{subfigure}{0.3\textwidth}
        \centering
        \includegraphics[width=\linewidth]{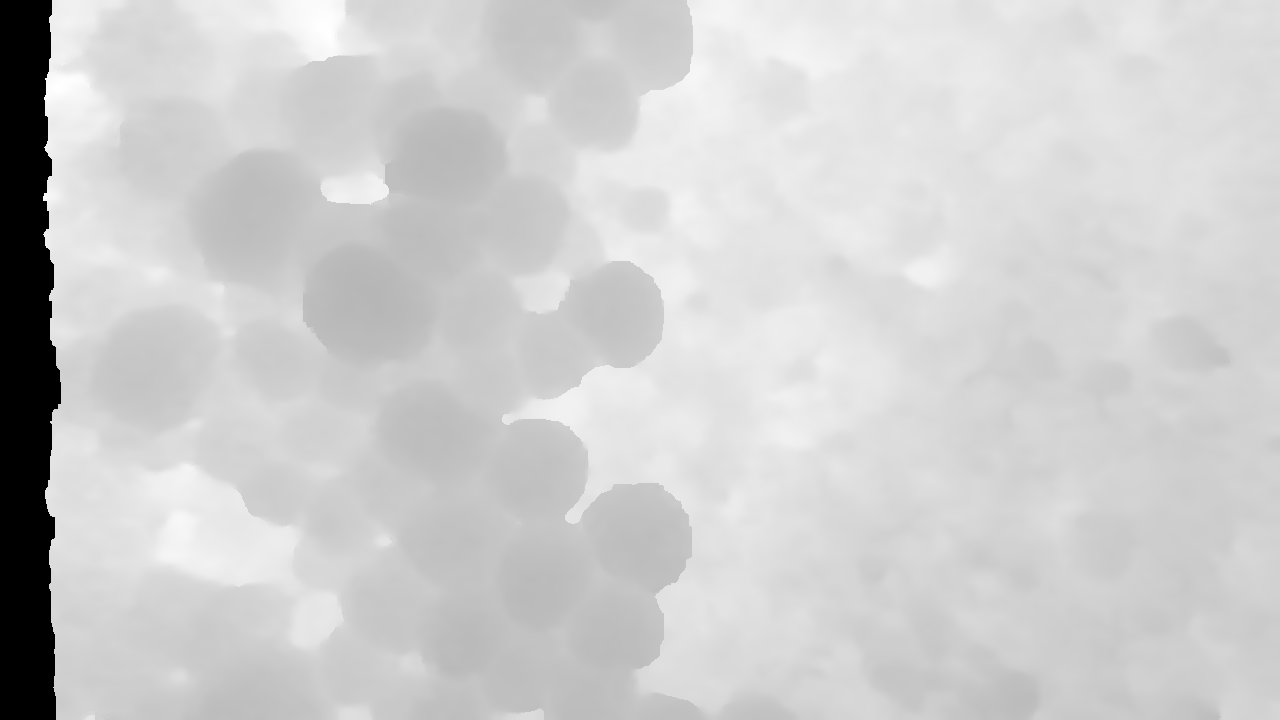}
        \caption{BB Sample Depth}
        \label{fig:sube}
    \end{subfigure}

    \begin{subfigure}{0.3\textwidth}
        \centering
        \includegraphics[width=\linewidth]{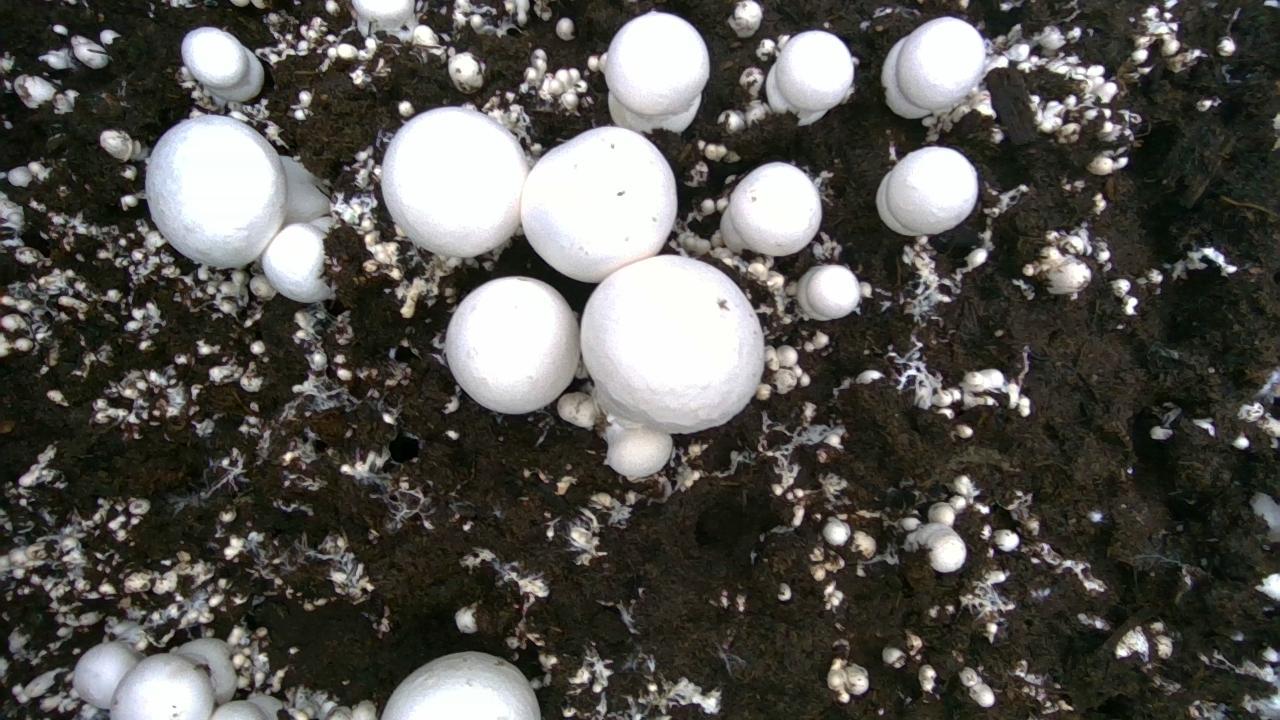}
        \caption{WB Sample Image}
        \label{fig:subb}
    \end{subfigure}
    \hfill
    \begin{subfigure}{0.3\textwidth}
        \centering
        \includegraphics[width=\linewidth]{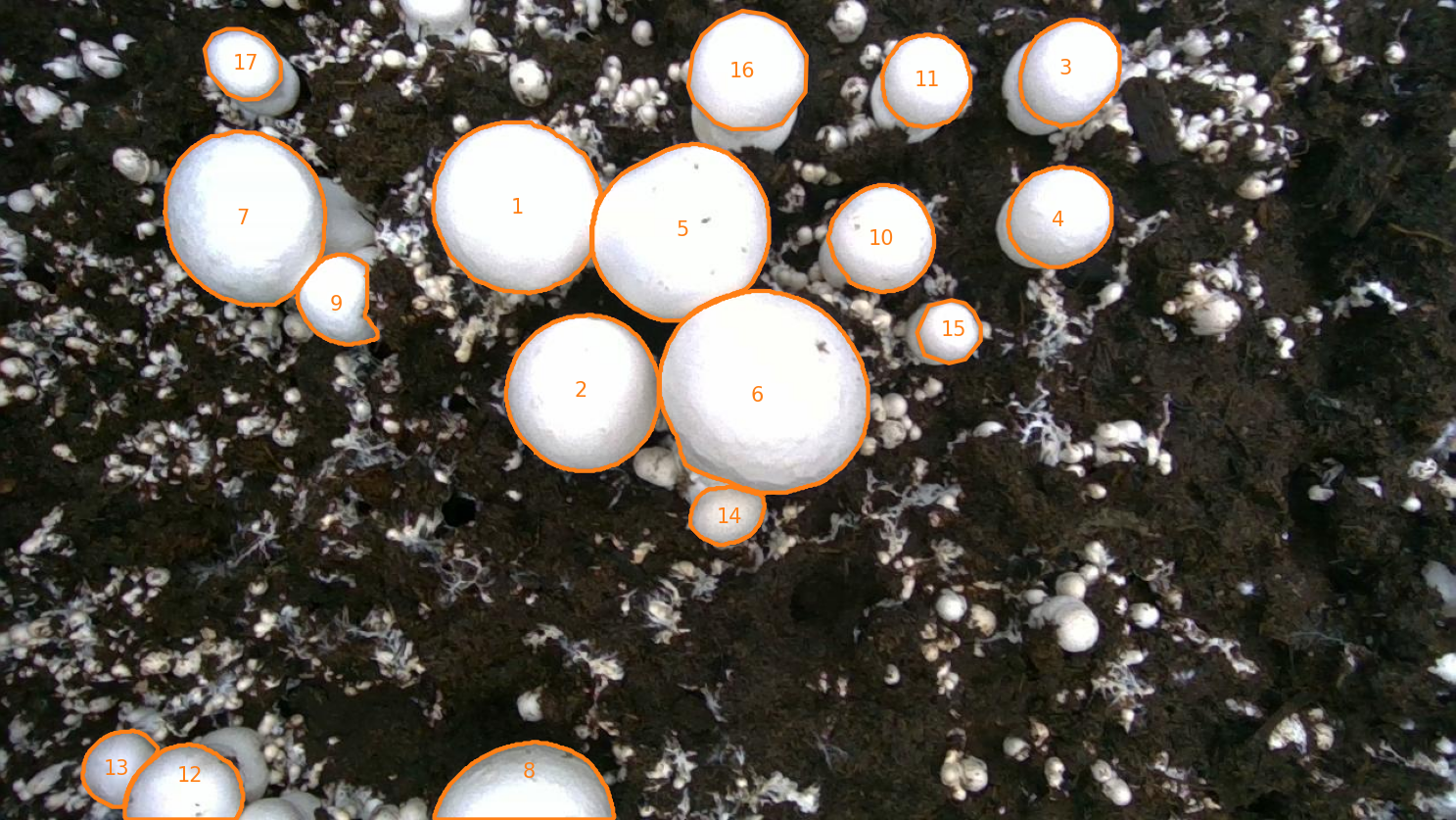}
        \caption{WB Sample Ground Truth}
        \label{fig:subd}
    \end{subfigure}
    \hfill
    \begin{subfigure}{0.31\textwidth}
        \centering
        \includegraphics[width=\linewidth]{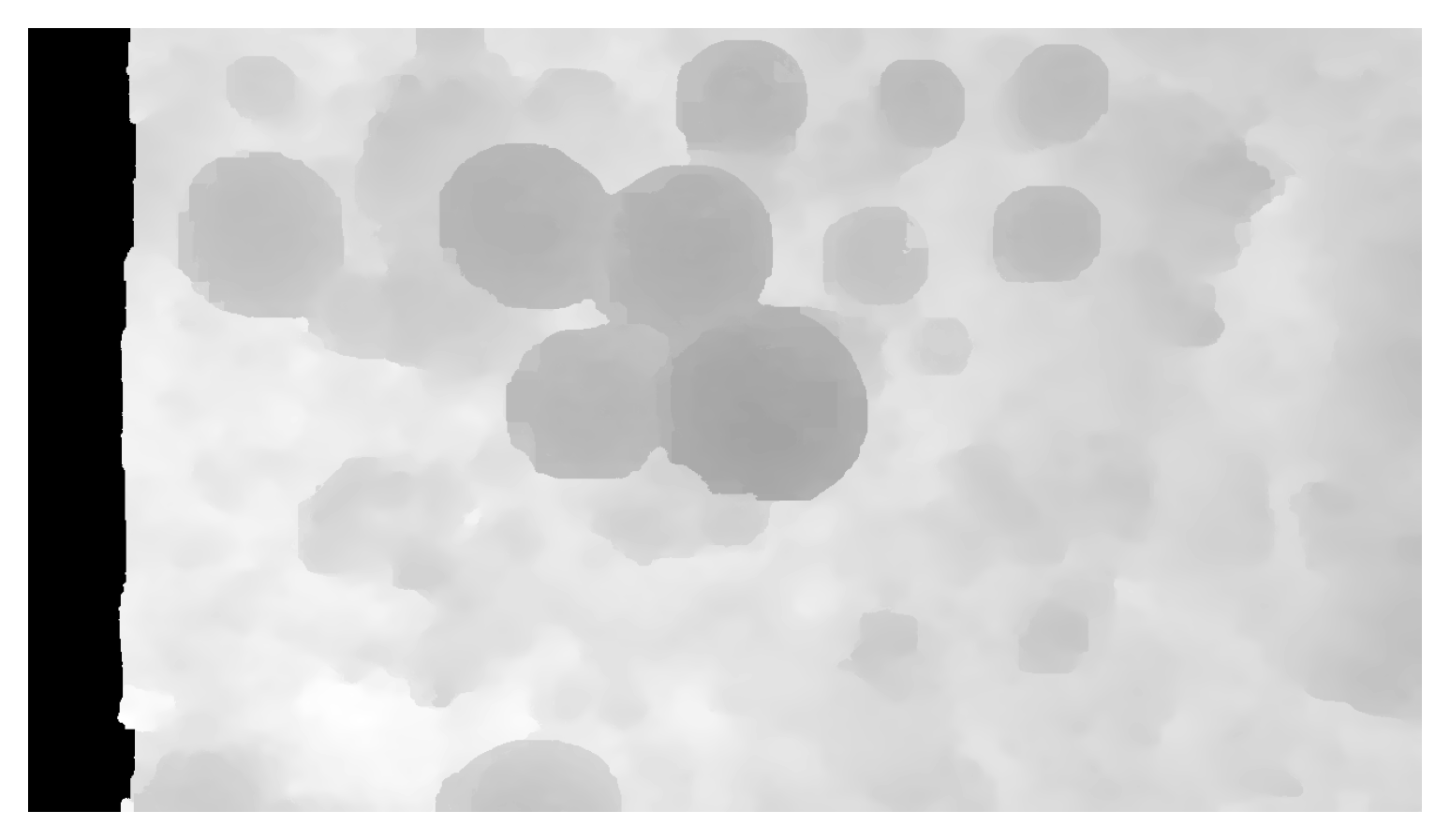}
        \caption{WB Sample Depth}
        \label{fig:subf}
    \end{subfigure}

    \caption{Sample of dataset images, ground truth instance segmentation masks, and depth images for Baby Bella (BB) and White Button (WB) mushroom images}
    \label{fig:full}
\end{figure*}

The dataset comprises two varieties of mushrooms commonly encountered in large-scale cultivation: white button mushrooms and baby bella (cremini) mushrooms. It consists of 173 RGB and depth image pairs of white button mushrooms, containing a total of 4,217 annotated instances, and 250 RGB and depth image pairs of baby bella mushrooms, containing a total of 13,955 annotated instances. The significant number of instances and images for each mushroom type provides a diverse and comprehensive dataset for benchmarking and evaluation.

Each image in the dataset encompasses various lighting conditions, capturing mushrooms at different growth stages, and encompasses a range of backgrounds and orientations. Accompanying the RGB images, the depth maps provide valuable depth information, allowing researchers to explore depth-based mushroom detection techniques for improved accuracy and robustness. The dataset further includes precise annotations for each image, providing detailed bounding boxes and segmentation mask annotations for individual mushrooms.

The creation of this mushroom dataset serves multiple objectives. First, it facilitates evaluating and comparing existing mushroom detection algorithms, enabling researchers to identify their strengths, weaknesses, and limitations. Second, it establishes a standardized benchmark for assessing the performance of novel mushroom detection methods, fostering the development of advanced mushroom detection methods for various tasks such as automatic mushroom harvesting, mushroom growth monitoring, and quality and quantity measurements.

Furthermore, the availability of this dataset addresses the lack of publicly accessible mushroom datasets, encouraging collaboration and promoting innovation in the field. Providing a shared platform for researchers to evaluate their techniques, encourages reproducibility and accelerates progress in mushroom detection technologies through the collective exchange of knowledge and resources.

\begin{figure}
    \centering
    \includegraphics[width=0.5\linewidth]{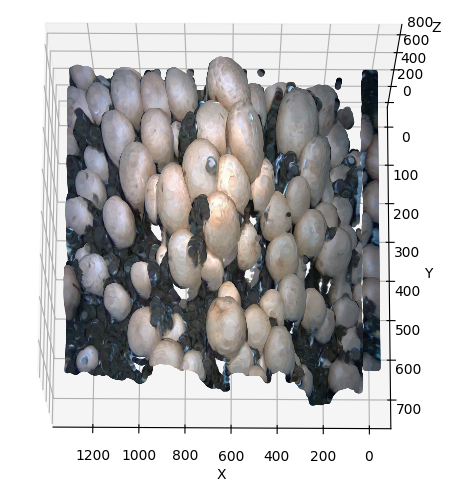}
    \caption{RGB-D Point Cloud Sample}
    \label{fig:point_cloud}
\end{figure}

In the subsequent sections of this paper, we provide a detailed description of the dataset, covering the annotation methodology, and comprehensive descriptive statistics. Additionally,  we provide an in-depth analysis of the dataset, highlighting its unique characteristics and potential challenges. Furthermore, we demonstrate the utility of the dataset by evaluating state-of-the-art object detection algorithms and discussing their performance within the context of our benchmark.
The introduction of the dedicated mushroom dataset in this study will play a pivotal role in the advancement of automatic mushroom harvesting systems. Our dataset, consisting of 423 RGB and depth image pairs, each with a resolution of 1280x720 pixels, along with precise annotations, provides a diverse and comprehensive resource for researchers to evaluate, compare, and develop mushroom detection methods. A sample of the dataset images, ground truth instance segmentation masks, and depth images are plotted in Fig. \ref{fig:full}. We anticipate that this dataset will significantly contribute to the progress and innovation in the field of edible mushroom smart and sustainable farming. 

\section{Literature Review}
Several articles and papers have been published regarding the application of image processing and machine learning algorithms for mushroom detection, quality assessment, disease detection, growth monitoring, and automatic harvesting. These works can be grouped into three main categories as follows.

\begin{table*}
\centering
\caption{
\label{table:works}
Summary of task, method, and results of similar studies leveraging mushroom detection. None of the datasets used in similar studies are publicly available. 
}

\scriptsize

\begin{tabularx}{\textwidth}{XXXXX}
\hline
Author            & Task                      & Method                                                   & Dataset Size        & Results \\ \hline
\cite{baisa_mushrooms_2022} & Detection, localization, and 3D pose estimation & Segmentation using AC, Detection using CHT, 3D localization using depth information & -                   & 98.99\% Precision 99.29\%  Recall \\ \hline
\cite{jareanpon_developing_2024}  & Fungal disease detection  & DenseNet201, ResNet50, Inception V3, VGGNet19            & 2000 images         & 94.35\% Precision 89.47\% F1-score \\ \hline
\cite{lee_development_2019}      & Detection, and maturity classification & Faster R-CNN for detection, SVM for maturity classification & 920 time-lapse image sets & 42.00\% Precision 82.00\% Recall 56.00\% F1-score 70.93\% Maturity classification accuracy \\ \hline
\cite{moysiadis_monitoring_2023} & Mushroom growth monitoring & YOLOv5 and Detectron2                                    & 1128 images, 4271 Mushrooms & 76.50\% F-1 Score 70.00\% Accuracy \\ \hline
\cite{nadim_application_2019}     & Mushroom quality control   & Neural network and fuzzy logic                           & 250 images          & 95.60\% Accuracy \\ \hline
\cite{olpin_region-based_2018}     & Detection                  & RCNN and RFCN                                            & 310 images          & 92.16\% Accuracy \\ \hline
\cite{retsinas_mushroom_2023}  & Detection and 3D pose estimation & segmentation using a k-Medoids approach based on FPFH and FCGF & Synthetic 3D dataset & 99.80\% MAP at 25.00\% IOU \\ \hline
 Vizhanyo and Felfoldi \cite{vizhanyo_enhancing_2000} & Disease Detection & LDA & - & 85.00\% True Classification Rate \\ \hline
\cite{wang_automatic_2018} & Automatic sorting & Watershed, Canny, Morphology & - & 97.42\% Accuracy \\ \hline 
\cite{wang_mushroom-yolo_2022} & Detection and growth monitoring & YOLOv5 + CBAM + BiFPN & - & 99.24\% MAP  \\ \hline 
\cite{wei_recursive-yolov5_2022} & Detection and growth monitoring & YOLOv5 + ASPP + CIOU & - & 98.00\% Accuracy  \\ \hline 
\cite{yang_research_2022} & Detection & MaskRCNN & - & 95.06\% AP at 50.00\% IOU  \\ \hline 
\cite{zahan_chapter_2022} & Disease Classification & AlexNet, GoogleNet, ResNet15 & 2536 images & 90.00\% Accuracy 89.00\% Precision 90.00\% Recall 90.00\% F1-score \\ \hline 

\end{tabularx}

\end{table*}

\subsection{Mushroom Detection and Localization}
There is a growing body of literature dedicated to mushroom detection and localization, each focusing on different algorithms for mushroom detection and localization, using techniques from classical image processing to advanced deep learning and computer vision.

The YOLO (You Only Look Once) algorithm, a significant advancement introduced in 2016, is a cornerstone in this field. It simplified object detection into a single regression problem, directly predicting bounding boxes and class probabilities, thus enhancing efficiency in detection tasks. Several studies adapted the YOLO algorithm for mushroom detection, yielding notable results.

\cite{wei_recursive-yolov5_2022} developed Recursive-YOLOv5, significantly surpassing the original YOLOv5 with 98\% accuracy, though at the cost of almost doubling the network's parameters.
\cite{wang_mushroom-yolo_2022} introduced Mushroom-YOLO, an improved version of YOLOv5, achieving a mean average precision of 99.24\% by integrating Convolutional Block Attention Module (CBAM) and Bidirectional Feature Pyramid Network (BiFPN) into the architecture.
\cite{olpin_region-based_2018} used Region-Based Convolutional Networks for mushroom detection, finding that the Region-based Convolutional Neural Network(RCNN) model was more accurate (92.162\%) than the Region-based Fully Convolutional Network (RFCN) (87.631\%), albeit slower.
\cite{yang_research_2022} enhanced Agaricus Bisporus recognition using Mask RCNN, achieving an Average Precision (AP50) of 95.061\%, though the operation was slow.
\cite{retsinas_mushroom_2023} created a vision module for 3D pose estimation of mushrooms, resulting in high retrieval scores (up to 99.80\% Mean Average Precision at 25\% Intersection over Union) and accurate pose estimation (mean angle error as low as 8.70°).
\cite{lee_development_2019} automated the selective harvesting process, using the Faster R-CNN model for identification and a 3D point cloud for segmentation, achieving 70.93\% accuracy in maturity identification.
Baisa and Al-Diri \cite{baisa_mushrooms_2022} focused on robotic picking, detecting, and estimating the 3D pose of mushrooms with high precision (98.99\%) and recall (99.29\%), using RGB-D data.
These studies demonstrate advancements in mushroom detection and localization, showcasing various algorithmic improvements. While they present significant progress, challenges such as computational efficiency, dataset quality, and the need for further accuracy and speed enhancements remain.

\subsection{Mushroom Growth Monitoring And Quality Assessment}

\cite{lu_development_2019} developed a system using YOLOv3 for monitoring mushroom growth in greenhouses. The system could estimate mushroom cap size, count, growth rate, and predict harvest times. It demonstrated effectiveness in identifying growth patterns and provided real-time updates via a mobile interface.
Lu and Liaw \cite{lu_novel_2020} introduced an image measurement algorithm utilizing YOLOv3, coupled with a Score-Punishment algorithm for accurate mushroom cap diameter measurement. The method excelled in accuracy over traditional methods but faced challenges with unclear mushroom contours or soil particles on caps.
\cite{nadim_application_2019} proposed an image processing system which assessed mushroom quality based on color, area, weight, and volume, using data mining, neural networks, and fuzzy logic. It achieved a 95.6\% correct detection rate but required image pre-processing to counteract quality issues from normal imaging conditions.

\cite{moysiadis_monitoring_2023} automated the monitoring and classification of oyster mushrooms in greenhouses. They used YOLOv5 for detection and classification, and Detectron2 for tracking growth, showing potential improvements in harvesting times but faced difficulties in detecting small mushrooms.
\cite{wang_automatic_2018} developed an automatic sorting system for white button mushrooms, using image processing to measure the pileus diameter. The system achieved high grading speed and accuracy, significantly improving over manual grading methods.

\cite{benhaddou} implemented a computer vision system to estimate mushroom yield and quality in a commercial farm setting, utilizing the Circular Hough Transform algorithm. They achieved detailed tracking of mushroom growth across 1960 frames over 20 days, demonstrating the system's ability to monitor size distribution and development trends. This technological application offers a promising tool for optimizing cultivation and harvesting strategies by providing precise data on mushroom growth patterns and potential yield estimations.

\subsection{Mushroom Disease Detection}

\cite{zahan_chapter_2022} employed deep learning models to classify mushroom diseases, with ResNet15 showing the best performance in accuracy and precision. The study emphasized the potential of deep learning in agricultural management.
Vizhanyo and Felfoldi \cite{vizhanyo_enhancing_2000} used advanced image analysis to distinguish diseased spots on mushroom caps. Their method showed an 85\% correct classification ratio, effectively differentiating between brown blotch and ginger blotch diseases.

\cite{jareanpon_developing_2024} developed an intelligent farm system for real-time detection of fungal diseases in Lentinus. The system integrated environment control, an imaging robot, and a deep learning-based prognosis system, achieving high precision in maintaining optimal conditions and disease detection.

In conclusion, the literature review section of this paper highlights a range of innovative approaches to mushroom detection, quality assessment, growth monitoring, and disease detection using image processing and machine learning algorithms. Table \ref{table:works} lists these studies and their results. The studies vary in focus, ranging from detection and localization to quality and disease detection, each contributing valuable insights and advancements in the field. Notably, while these works demonstrate significant progress, there remains a gap in the standardization of methodologies and results comparison, compounded with challenge of data drift \cite{Mirza_2024_DQA}.  This gap is primarily due to the lack of a publicly available benchmark dataset. Based on the current literature \cite{yin_computer_2022}, and to the best of the authors' knowledge, there are no publicly available annotated edible mushroom datasets. A standardized dataset would provide a common ground for evaluating different algorithms and techniques, ensuring comparability and consistency across studies. The availability of such a dataset would benefit the field, allowing for more robust and reliable comparisons of methods and results, driving forward the development of more efficient and effective solutions for mushroom cultivation and processing in the agricultural sector.   

\section{Labeling Process}
The images were initially labeled using the Segment Anything Model (SAM) automatic mask generation feature \cite{kirillov2023segment}. This feature provides several configurations and post-processing steps to ensure the best quality of automatically generated masks for specific purposes. For instance, a grid of n*n (n=32 by default) positive points is placed on the original image and randomly cropped versions of it, and the generated masks are filtered by their confidence scores and stability within different confidence thresholds. The automatic mask generation algorithm considers a mask to be stable if thresholding the probability map at \(0.5 - \delta\) and \(0.5 + \delta\) results in similar masks. Additionally, masks can be filtered by their minimum and maximum area and a certain maximum intersection over union (IOU) threshold can be set to avoid generation of overlapping masks. Nevertheless, these configurations do not guarantee the generation of perfect ground truth masks and can only be used as a starting point. These generated masks were meticulously edited manually during several steps to guarantee the perfection of the dataset ground truth masks. Initially, all the false positive masks were removed and the missing masks were added. Later, the intersection of all pairs of masks in the same image was calculated for all the images in the dataset, and the overlapping masks were removed manually.
   
\section{Data Description}
Our dataset includes 423 RGBD images, 250 of which belong to Baby Bella mushrooms and the remaining 173 belong to the White Button mushrooms. A sample of the dataset images, depth maps, and label masks is plotted in Fig.\ref{fig:full}. Both the RGB images and the depth images have a resolution of 1280*720 pixels, and hence, no alignment will be required for using the RGB and depth images as a single RGB-D input. A sample 3D point cloud image is plotted in Fig. \ref{fig:point_cloud}. The camera was placed at vertical distances of 27 cm (about 10.63 in) and 15 cm (about 5.91 in) above the cultivation beds during image acquisition of baby bella and white button mushrooms, respectively.

\begin{figure}
    \centering
    \includegraphics[width=\textwidth]{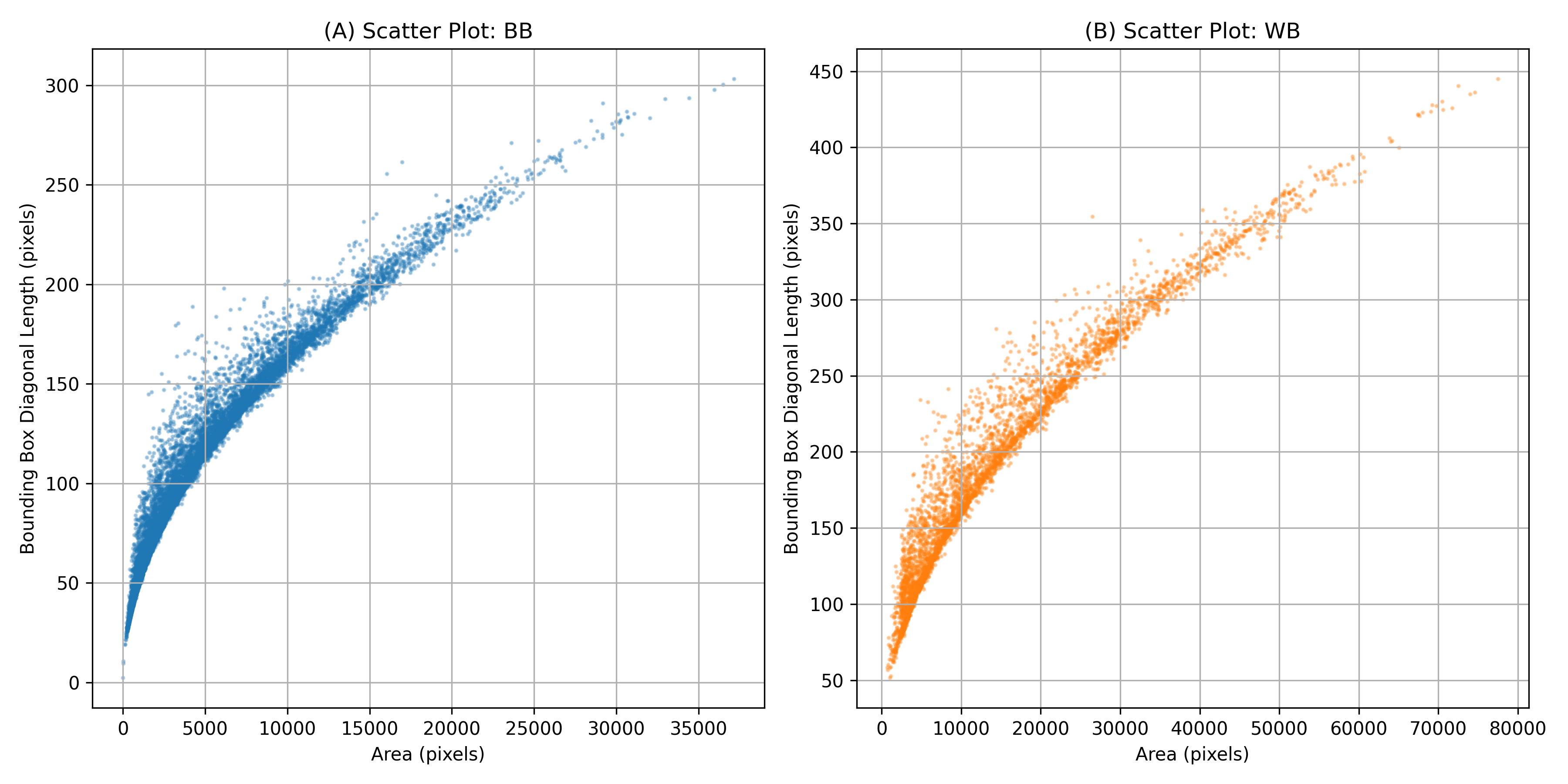}
    \caption{Scatter plots of instances mask area vs bounding box diagonal length for (A) baby bella mushrooms and (B) white button mushrooms.}
    \label{fig:scatter}
\end{figure}

\begin{figure}
    \centering
    \includegraphics[width=0.80\linewidth]{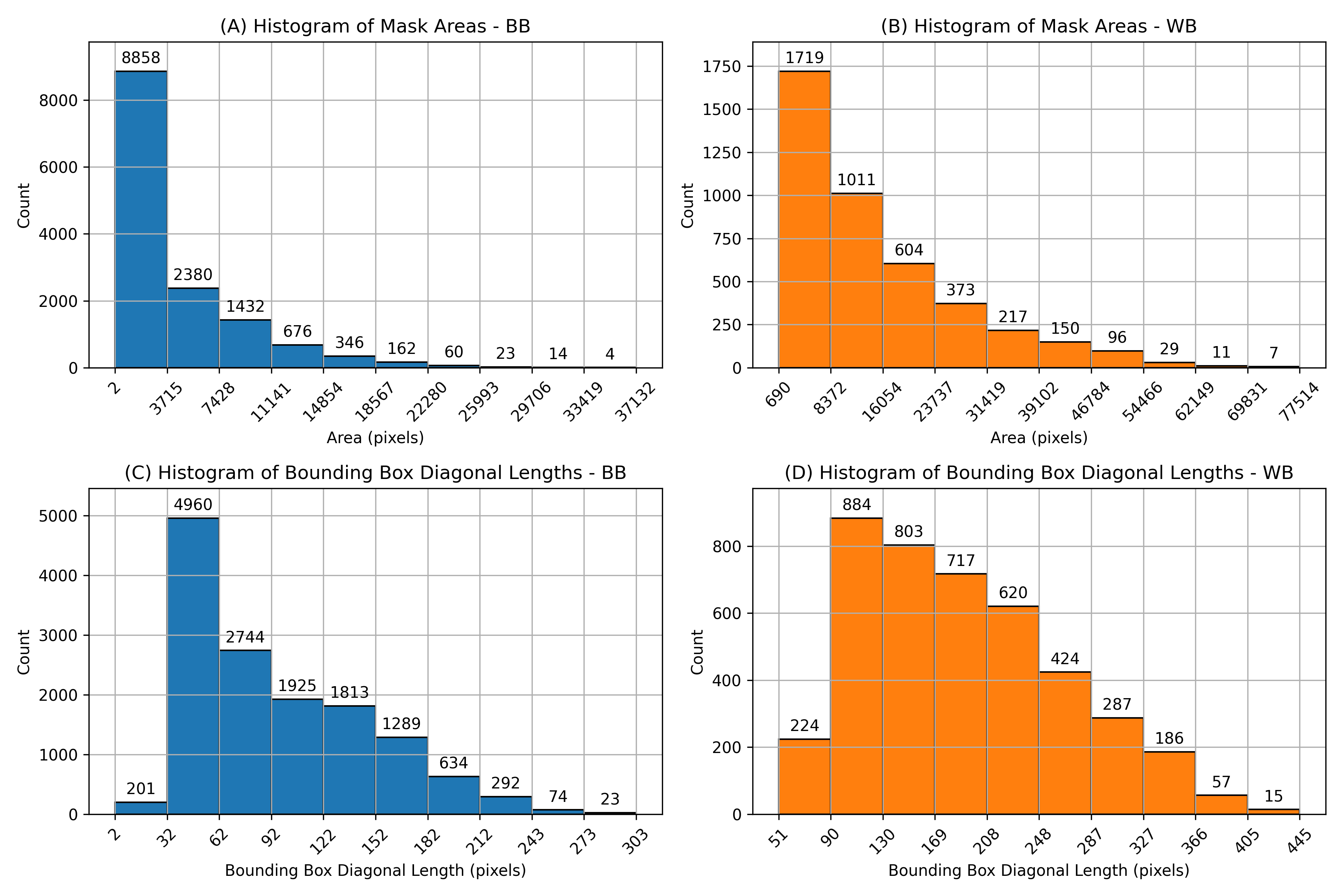}
    \caption{Histograms of Mask Areas and Bounding Box Diagonal Lengths; Histograms (A) and (B) show pixel mask areas of Baby Bella (BB) and White Button (WB) mushrooms accordingly. Histograms (C) and (D) show bounding box diagonal lengths for BB and WB mushroom instances accordingly.}
    \label{fig:histograms}
\end{figure}

Our dataset consists of 4217 white button mushroom instances and 13,955 baby bella mushroom instances in total. These instances include a wide range of mushroom sizes for both mushroom categories which makes our dataset suitable for training detection models aiming for various tasks such as mushroom growth monitoring and mushroom automatic harvesting. Histograms of mask areas and bounding box diagonal length of mushroom instances are plotted in Fig. \ref{fig:histograms} and scatter plots of instances mask area vs bounding box diagonal length are plotted in Fig. \ref{fig:scatter}. As can be seen in the scatter plots, a fraction of the mushroom instances are above the scatter plot’s trend line, meaning they have a small area compared to their diagonal length. These instances usually belong to the mushrooms occluded by neighboring mushrooms in the cluster, causing only a portion of them to be visible, having crescent-shaped masks as opposed to fully visible mushrooms that have convex-shaped masks. Both figures \ref{fig:scatter} and \ref{fig:histograms} show much larger masks for white button mushrooms which is caused by the smaller vertical distance between the camera and cultivation bed at the time of image acquisition.

\begin{figure}[!h]
    \centering

    \begin{subfigure}{0.33\textwidth}
        \centering
        \includegraphics[width=\textwidth]{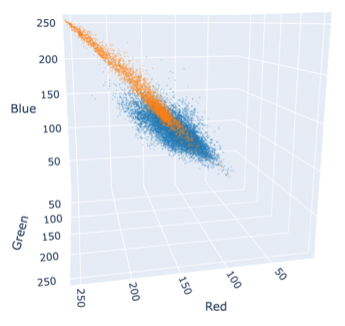}
        \caption{RGB}
        \label{subfig:image1}
    \end{subfigure}%
    \hfill
    \begin{subfigure}{0.33\textwidth}
        \centering
        \includegraphics[width=\textwidth]{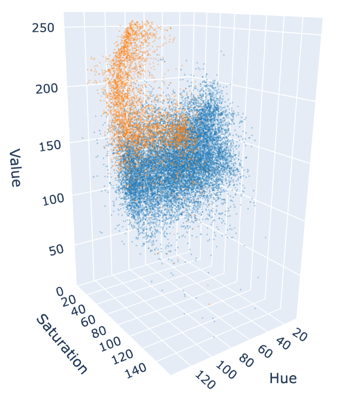}
        \caption{HSV}
        \label{subfig:image2}
    \end{subfigure}%
    \hfill
    \begin{subfigure}{0.33\textwidth}
        \centering
        \includegraphics[width=\textwidth]{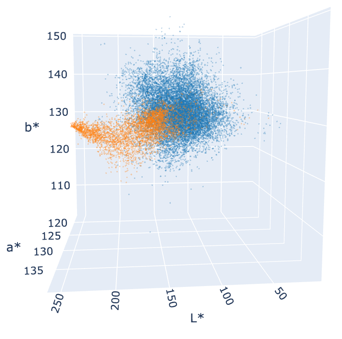}
        \caption{LAB}
        \label{subfig:image3}
    \end{subfigure}

    \caption{Color distribution of mushroom instances in various color spaces; Judging only by the color distributions, LAB and HSV color spaces are more suitable than RGB for classifying WB and BB instances. WB mushrooms are shown in orange and BB mushrooms are shown in blue.}
    \label{fig:three_images}
\end{figure}

To study the separability of the two mushroom classes under various color spaces, we have plotted scatter plots visualizing the distribution of instances of each class in three color spaces RGB, HSV, and LAB in Fig. \ref{fig:three_images}. Each data point on the scatter plots represents a single instance and the mean value of all the pixels of the instance's mask for each color channel were translated as x, y, and z coordinates of the 3D scatter plots. The two clusters seem more intertwined in RGB color space compared to HSV and LAB. 

\section{Benchmarking and Results}

In this section, we provide a detailed analysis of several detection and instance segmentation algorithms tested on our mushroom dataset. The models were rigorously trained on a robust hardware setup consisting of a DGX-2 cluster of 16 Nvidia V-100 GPUs, each equipped with 32GB of memory, ensuring substantial computational power for handling extensive data and complex model architectures. All models underwent thorough evaluations, with their performance presented in Table \ref{results}. Our assessment utilized metrics such as F1 score, average precision (AP), and average recall (AR) for all models regardless of the task being detection or instance segmentation. 

\begin{figure}[!ht]
  \centering
  \label{fig:visualization-a}
  \begin{subfigure}{0.48\textwidth}
    \centering
    \includegraphics[width=\linewidth]{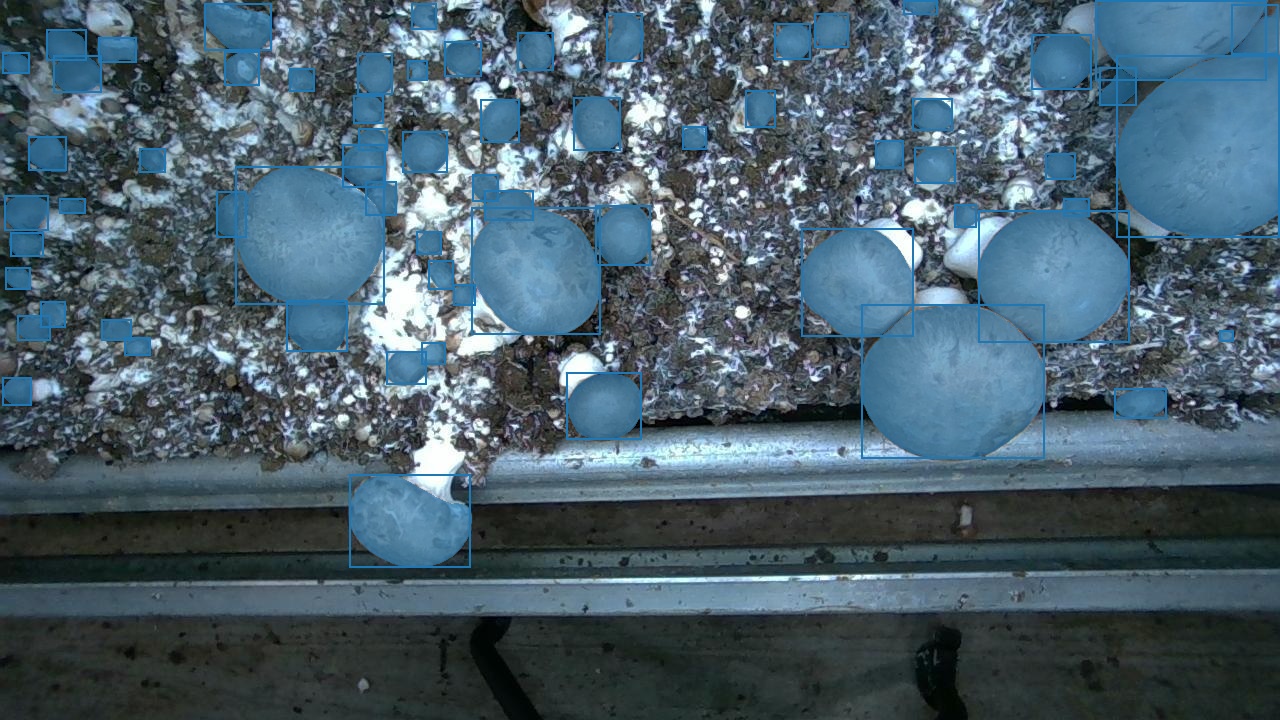}
    \caption{BB Sample Image}
  \end{subfigure}
  \hfill
  \begin{subfigure}{0.48\textwidth}
    \label{fig:visualization-b}
    \centering
    \includegraphics[width=\linewidth]{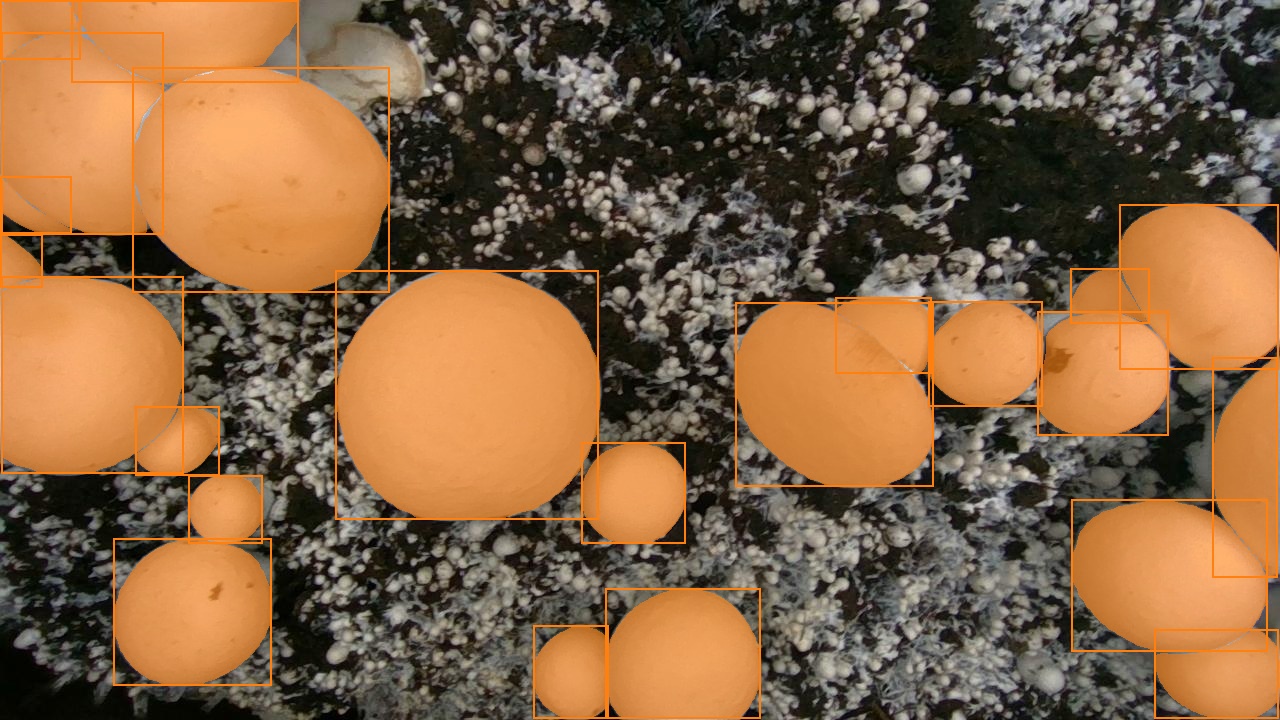}
    \caption{WB Sample Image}
  \end{subfigure}
  \caption{Prediction results of MaskRCNN-ResNet-50 architecture on (a) baby bella images, and (b) white button mushroom images}
  \label{fig:visualization}
\end{figure}

\begin{table*}
  \centering
  \caption{Performance of different object detection and instance segmentation models on M18K dataset - Params is number of model parameters in Million, and C is number of input channels}
  \label{results}
  {\scriptsize
  \begin{tblr}{
    width = \linewidth,
    colspec = {X[0.8]X[1]X[0.4]X[0.3]X[0.3]X[0.3]X[0.3]X[0.3]X[0.3]},
    rowsep = 0.5pt,
    cell{2}{1} = {r=3}{},
    cell{5}{1} = {r=8}{},
    cell{5}{2} = {r=2}{},
    cell{5}{3} = {r=2}{},
    cell{5}{4} = {r=2}{},
    cell{7}{2} = {r=2}{},
    cell{7}{3} = {r=2}{},
    cell{7}{4} = {r=2}{},
    cell{9}{2} = {r=2}{},
    cell{9}{3} = {r=2}{},
    cell{9}{4} = {r=2}{},
    cell{11}{2} = {r=2}{},
    cell{11}{3} = {r=2}{},
    cell{11}{4} = {r=2}{},
    cell{13}{1} = {r=2}{},
    cell{13}{2} = {r=2}{},
    cell{13}{3} = {r=2}{},
    cell{13}{4} = {r=2}{},
    cell{13}{5} = {r=2}{},
    cell{15}{1} = {r=2}{},
    cell{15}{2} = {r=2}{},
    cell{15}{3} = {r=2}{},
    cell{15}{4} = {r=2}{},
    cell{15}{5} = {r=2}{},
    cell{17}{1} = {r=4}{},
    cell{17}{2} = {r=4}{},
    cell{17}{3} = {r=4}{},
    cell{17}{4} = {r=4}{},
    cell{17}{5} = {r=2}{},
    cell{19}{5} = {r=2}{},
    cell{21}{1} = {r=4}{},
    cell{21}{2} = {r=4}{},
    cell{21}{3} = {r=4}{},
    cell{21}{4} = {r=4}{},
    cell{21}{5} = {r=2}{},
    cell{23}{5} = {r=2}{},
    cell{25}{1} = {r=4}{},
    cell{25}{2} = {r=4}{},
    cell{25}{3} = {r=4}{},
    cell{25}{4} = {r=4}{},
    cell{25}{5} = {r=2}{},
    cell{27}{5} = {r=2}{},
    cell{29}{1} = {r=4}{},
    cell{29}{2} = {r=4}{},
    cell{29}{3} = {r=4}{},
    cell{29}{4} = {r=4}{},
    cell{29}{5} = {r=2}{},
    cell{31}{5} = {r=2}{},
    cell{33}{1} = {r=4}{},
    cell{33}{2} = {r=4}{},
    cell{33}{3} = {r=4}{},
    cell{33}{4} = {r=4}{},
    cell{33}{5} = {r=2}{},
    cell{35}{5} = {r=2}{},
    cell{37}{1} = {r=2}{},
    cell{37}{2} = {r=2}{},
    cell{37}{3} = {r=2}{},
    cell{37}{4} = {r=2}{},
    cell{37}{5} = {r=2}{},
    cell{39}{1} = {r=2}{},
    cell{39}{2} = {r=2}{},
    cell{39}{3} = {r=2}{},
    cell{39}{4} = {r=2}{},
    cell{39}{5} = {r=2}{},
    hline{1,2,5,13,17,21,25,29,33,37,39,41} = {-}{},
    hline{3-4,7,9,11} = {2-9}{},
    hline{6,8,10,12,19,23,27,31,35} = {5-9}{},
    hline{14,16,18,20,22,24,26,28,30,32,34,36,38,40} = {6-9}{},
    hline{15} = {1-2,3-9}{},
  }
  \textbf{Model Name} & \textbf{Backbone}         & \textbf{Params} & \textbf{C} & \textbf{Task}         & \textbf{Size} & \textbf{F1}   & \textbf{AP} & \textbf{AR} \\
  FasterRCNN \cite{ren_faster_2015}  & efficientnet\_b1 & 86.6   & 3        & Det    & 1280        & 0.824       & 0.807              & 0.842           \\
              & MobileNet-V3 \cite{Howard_2019_MobileNetV3}    & 18.9   & 3        & Det    & 1280        & 0.825       & 0.808              & 0.843           \\
              & ResNet-50 \cite{He_2016_ResNet}       & 43.3   & 3        & Det    & 1280        & 0.894       & 0.880              & 0.909           \\
  MaskRCNN \cite{He_2017_MaskRCNN}    & EfficientNet-B1 \cite{tan2020efficientnet} & 91.6   & 3        & Det    & 1280        & 0.819       & 0.802              & 0.836           \\
              &                  &        &          & Seg & 1280        & 0.812       & 0.797               & 0.828           \\
              & MobileNet-V3    & 64.8   & 3        & Det    & 1280        & 0.816       & 0.800              & 0.834            \\
              &                  &        &          & Seg & 1280        & 0.813       & 0.798              & 0.828           \\
              & ResNet-50       & 45.9   & 3        & Det    & 1280        & 0.891 & 0.876              & 0.905           \\
              &                  &        &          & Seg & 1280        & \textbf{0.881} & \textbf{0.866}              & \textbf{0.896}            \\
              & ResNet-50 & 171    & 4        & Det    & 1280        & 0.863 & 0.846              & 0.881           \\
              &                  &        &          & Seg & 1280        & 0.861  & 0.846              & 0.876           \\
  RT-DETR-L \cite{lv2023detrs}    & ResNet-50       & 31.9   & 3        & Det    & 640         & 0.245  & 0.210               & 0.295           \\
              &                  &        &          &              & 1280        & \textbf{0.907} & \textbf{0.891}              & \textbf{0.923}           \\
  RT-DETR-X    & ResNet-50       & 65.4   & 3        & Det    & 640         & 0.120 & 0.099              & 0.153           \\
              &                  &        &          &              & 1280        & 0.906  & \textbf{0.891}              & 0.922           \\
  YOLOV8-L & CSPDarknet \cite{Wang_2020_cspnet}      & 43.7   & 3        & Det    & 640         & 0.735 & 0.693              & 0.782           \\
              &                  &        &          &              & 1280        & 0.880 & 0.869              & 0.892           \\
              &                  &        &          & Seg & 640         & 0.700 & 0.659              & 0.745           \\
              &                  &        &          &              & 1280        & 0.857 & 0.846              & 0.868            \\
  YOLOV8-M & CSPDarknet       & 25.9   & 3        & Det    & 640         & 0.732 & 0.690              & 0.779            \\
              &                  &        &          &              & 1280        & 0.868 & 0.857               & 0.880           \\
              &                  &        &          & Seg & 640         & 0.711 & 0.669              & 0.760           \\
              &                  &        &          &              & 1280        & 0.844 & 0.833              & 0.856           \\
  YOLOV8-N & CSPDarknet       & 3.2    & 3        & Det    & 640         & 0.775  & 0.739              & 0.815           \\
              &                  &        &          &              & 1280        & 0.887 & 0.873              & 0.901           \\
              &                  &        &          & Seg & 640         & 0.748 & 0.712              & 0.788           \\
              &                  &        &          &              & 1280        & 0.859  & 0.846              & 0.874             \\
  YOLOV8-S & CSPDarknet       & 11.2   & 3        & Det    & 640         & 0.775 & 0.743              & 0.809            \\
              &                  &        &          &              & 1280        & 0.880 & 0.868               & 0.892           \\
              &                  &        &          & Seg & 640         & 0.753 & 0.720              & 0.789           \\
              &                  &        &          &              & 1280        & 0.857 & 0.845              & 0.869           \\
  YOLOV8-X & CSPDarknet       & 68.3   & 3        & Det    & 640         & 0.724 & 0.667               & 0.790           \\
              &                  &        &          &              & 1280        & 0.887 & 0.875              & 0.898           \\
              &                  &        &          & Seg & 640         & 0.695 & 0.638              & 0.763           \\
              &                  &        &          &              & 1280        & 0.858 & 0.847              & 0.869           \\
  YOLOV9-C \cite{wang2024yolov9}     & GELAN-C          & 25.5   & 3        & Det    & 640         & 0.760  & 0.726              & 0.797           \\
              &                  &        &          &              & 1280        & 0.884 & 0.872              & 0.897           \\
  YOLOV9-E     & GELAN-E          & 58.1   & 3        & Det    & 640         & 0.736 & 0.702              & 0.774           \\
              &                  &        &          &              & 1280        & 0.877  & 0.867              & 0.887           
  \end{tblr}
  }
  \end{table*}

The data in Table \ref{results} provides details of the performance of a variety of models with distinct backbone architectures, benchmarked on object detection and instance segmentation tasks. These models were trained on image resolutions of \textbf{640} * 360 and \textbf{1280} * 720. The MaskRCNN with a ResNet-50 backbone surpassed the other networks in instance segmentation, achieving a mean Average Precision (mAP) of 0.866 and an Average Recall (AR) of 0.896 when processing 1280-sized RGB images. When adapted for RGB-D input, which increased the network's parameters over three times the original, there was no enhancement in performance. This lack of improvement may stem from the mandatory re-initialization of weights following the modifications to accommodate the added depth channel, which precluded the use of existing pre-trained weights that typically contribute to superior performance via fine-tuning on a specific dataset. Furthermore, the expanded parameter count makes the network less suitable for real-time use where hardware resources are constrained. Due to the lack of improvement in performance while adding the depth channel, other networks were not tested with RGB-D images. Furthermore, The RT-DETR Large model exhibited superior object detection performance on 1280-sized images, with a mAP of 0.907 and an AR of 0.923. In contrast, both RT-DETR variants performed poorly on 640-sized images, potentially due to the network not being as effective at capturing the necessary details at a reduced resolution given the small number of dataset images. An intriguing finding was that the smallest and largest YOLO-V8 versions, YOLOV8-Nano with 3.2 million parameters, and YOLOV8-X with 68.3 million parameters, achieved nearly identical mAPs of 0.873 and 0.875, respectively. This suggests that beyond a certain threshold, additional parameters may yield diminishing returns in performance improvement, and that smaller models may be sufficiently expressive for our given task or more effective in generalizing from the small number of training data. Fig. \ref{fig:visualization} provides a visual comparison of the prediction results for Mask R-CNN with a ResNet-50 backbone trained on RGB images of size 1280. Fig. \ref{fig:visualization}(a) and Fig. \ref{fig:visualization}(b) show the prediction results on RGB images for BB (Baby Bella) and WB (White Buttons) mushroom instances, respectively.
Overall, the results indicate that while all models performed well across various metrics, the choice of the backbone and the balance between the number of parameters and input channels play a significant role in the trade-off between accuracy and computational efficiency. These insights can inform the development and optimization of mushroom detection algorithms tailored for agricultural automation.

\section{Conclusion}
In this work, we have presented a dedicated mushroom detection dataset tailored to train and evaluate object detection and instance segmentation methods for various purposes that require localization of mushrooms in images such as automated mushroom harvesting, mushroom quality assessment, and disease detection. Through comprehensive data collection and meticulous labeling processes, the dataset encompasses a wide spectrum of scenarios and challenges inherent in real-world mushroom growth environments including clusters of mushrooms with a variety of densities, images with different lighting conditions, occasional occlusion of mushroom instances by soil particles, and mushrooms growing at the edge of cultivation beds. The diverse nature of the dataset, including instances of both white button and baby bella mushrooms, allows for the robust training and benchmarking of detection and instance segmentation algorithms.

Our evaluation of 11 different models highlights their respective strengths and weaknesses, shedding light on their suitability for mushroom detection and instance segmentation tasks. Furthermore, our exploration of color spaces provides insights into potential improvements in classification accuracy. By sharing this dataset and evaluation results, we hope to catalyze research and innovation in the domain of smart edible mushroom farming, fostering collaborations and accelerating progress in detection technologies. The availability of this benchmark dataset, our code, and trained models contributes to the development of advanced mushroom detection techniques, paving the way for increased efficiency and sustainability in automated mushroom harvesting systems.





\section*{Acknowledgements and funding}
This work is partially supported by the United States Department of Agriculture grants \#13111855, 13424031, and 2023-51300-40853, and the University of Houston Infrastructure Grant. The authors would also like to acknowledge Monterey Mushroom Farms, Madisonville, Texas for their collaboration on this study.

\bibliographystyle{elsarticle-harv} 
\bibliography{bib}

\end{document}